\documentclass{article}

    \usepackage[final]{neurips_2022}

\usepackage[utf8]{inputenc} 
\usepackage[T1]{fontenc}    
\usepackage{hyperref}       
\usepackage{url}            
\usepackage{booktabs}       
\usepackage{amsfonts}       
\usepackage{nicefrac}       
\usepackage{microtype}      
\usepackage{xcolor}         

\usepackage{pgfplots}
\usepgfplotslibrary{fillbetween}
\usetikzlibrary{patterns}
\usetikzlibrary{pgfplots.groupplots}
\pgfplotsset{compat=1.17}
\usepackage{lscape}
\usepackage{multirow}
\usepackage{dblfloatfix}
\usepackage{footnote}
\usepackage{graphicx}
\usepackage{tikz}
\usepackage{hyphenat}
\usepackage{float}
\usepackage{amsmath, mathtools}
\usepackage{amssymb}
\usepackage{booktabs, tabularx}
\usepackage{amssymb}
\usepackage{subfig}
\usepackage{wrapfig}

\title{BERT on a Data Diet: 
Finding Important Examples by Gradient-Based Pruning
}

\author{%
  Mohsen Fayyaz
  \thanks{Equal contribution.} \\
  University of Tehran\\
  \texttt{mohsen.fayyaz77@ut.ac.ir} \\
   \And
   Ehsan Aghazadeh$^{\ast}$ \\
   University of Tehran \\
   \texttt{eaghazade1998@ut.ac.ir} \\
   \AND
   Ali Modarressi$^{\ast}$ \\
   Iran University of Science and Technology \\
   \texttt{amodaresi@gmail.com} \\
   \And
   Mohammad Taher Pilehvar \\
   Tehran Institute for Advanced Studies \\
   Khatam University, Iran \\
   \texttt{mp792@cam.ac.uk} \\
   \And
   Yadollah Yaghoobzadeh \\
   University of Tehran \\
   \texttt{y.yaghoobzadeh@ut.ac.ir} \\
   \And
   Samira Ebrahimi Kahou \\
   ÉTS Montréal, Mila, CIFAR \\
   \texttt{samira.ebrahimi-kahou@etsmtl.ca} \\
}

\begin{document}

\maketitle

\begin{abstract}
Current pre-trained language models rely on large datasets for achieving state-of-the-art performance. 
However, past research has shown that not all examples in a dataset are equally important during training.
In fact, it is sometimes possible to prune a considerable fraction of the training set while maintaining the test performance. 
Established on standard vision benchmarks, two gradient-based scoring metrics for finding important examples are \emph{GraNd} and its estimated version, \emph{EL2N}. 
In this work, we employ these two metrics for the first time in NLP.
We demonstrate that these metrics need to be computed after at least one epoch of fine-tuning and they are not reliable in early steps. 
Furthermore, we show that by pruning a small portion of the examples with the highest GraNd/EL2N scores, we can not only preserve the test accuracy, but also surpass it. 
This paper details adjustments and implementation choices which enable \emph{GraNd} and \emph{EL2N} to be applied to NLP.
\end{abstract}

\section{Introduction}
Large datasets have made the phenomenal performance of Transformer-based \citep{attention-is-all-you-need} pre-trained language models (PLMs) possible \citep{devlin-etal-2019-bert, sanh2019distilbert, liu-2019-roberta, clark2020electra, bommasani2021fm}. 
However, recent studies show that a significant fraction of examples in a training dataset can be omitted without sacrificing test accuracy. 
To this end, many metrics have been introduced for ranking the examples in a dataset based on their importance \citep{toneva2018empirical-forgetting, swayamdipta-2020-dataset-cartography, yaghoobzadeh-etal-2021-increasing, varshney-mishra-and-chitta-baral-2022-model}. 
One of these metrics is \emph{Forgetting Score} \citep{toneva2018empirical-forgetting, yaghoobzadeh-etal-2021-increasing} which recognizes the examples that are misclassified after being correctly classified during training or are always misclassifed. 
\emph{Datamap} \citep{swayamdipta-2020-dataset-cartography} is another technique for diagnosing datasets which uses \emph{confidence} and \emph{variability} metrics. 

\emph{GraNd} and its approximation, \emph{EL2N}, are two recently introduced metrics that have only been studied for image classification models and tasks \citep{NEURIPS2021_ac56f8fe_datadiet}.
The goal of this study is to adapt these metrics for PLMs and apply them in NLP tasks.
We select a topic classification and a natural language inference dataset to run our experiments. 
We find that training with PLMs instead of randomly initialized models, used in \citet{NEURIPS2021_ac56f8fe_datadiet}, brings new challenges. 
Besides, adapting these metrics to NLP is non-trivial because of the inherent differences between the visual and language modalities.

In summary, our contributions are threefold:
(1) we adapt GraNd and EL2N metrics to the language domain to identify important examples in a dataset;
(2) we show that in contrary to the results in computer vision, early score computation steps are not sufficient for finding a proper subset of the data in NLP; and 
(3) we observe that pruning a small fraction of the examples with the highest EL2N/GraNd scores will result in better performance and in some cases even better than fine-tuning on the whole dataset.


\section{Background}
In this section, we describe the two metrics introduced by \citet{NEURIPS2021_ac56f8fe_datadiet} for pruning the training data: \textbf{GraNd} and its estimated variant, \textbf{EL2N}. 
\subsection{GraNd}
Consider $\mathbf{X} = \{x_i, y_i\}_{i=1}^N$ to be a training dataset for a given classification task with $K$ classes, where $x_i$ is the input (i.e. a sequence of tokens) and $y_i$ is its corresponding label. 
To estimate the importance of each training sample $(x_i, y_i)$,  \citet{NEURIPS2021_ac56f8fe_datadiet} propose utilizing the expected value of the loss gradient norm denoted as GraNd:
\begin{equation}
\label{eq:grand}
    \text{GraNd}(x_i, y_i) = \mathbb{E}_{\boldsymbol{w}}\left\|\boldsymbol{g}(x_i, y_i)\right\|_2
\end{equation}
The vector $\boldsymbol{g}$ is the loss gradient with respect to the model's weights.  
This method is based on the assumption that the expected impact of a sample on the network weights ($\boldsymbol{w}$) denotes the importance of that sample.
By sorting and ranking the training data, a top subset could be used for the training process, while pruning out the remaining data with lower scores.

Note that, unlike the original method stated by \citet{NEURIPS2021_ac56f8fe_datadiet}, which initializes the entire network with random weights, we start our training with a pre-trained language model, a common practice in current NLP. Therefore, we compute the GraNd scores only for the randomly initialized classifier layer on top of the PLM.

\subsection{EL2N}
By defining $\psi^{(k)}(x_i) = \nabla_{\boldsymbol{w}}f^{(k)}(x_i)$ for the $k^\text{th}$ logit,
the loss gradient ($\boldsymbol{g}$) can be written as:
\begin{equation}
    \boldsymbol{g}(x_i, y_i) = \sum_{k=1}^{K}\nabla_{f^{(k)}}\mathcal{L}(f(x_i), y_i)^T\psi^{(k)}(x_i)
\end{equation}
Since the $\mathcal{L}(f(x_i), y_i)$ is a cross entropy loss, $\nabla_{f^{(k)}}\mathcal{L}(f(x_i), y_i)^T=p(x_i)^{(k)} - y^{(k)}_i$ , where $p(x_i)$ is the output probability vector of the model. By assuming orthogonality and uniform sizes across logits for $\psi$ over the training examples, the norm in Eq. \ref{eq:grand} is approximately equal to $\left\|p(x_i) - \boldsymbol{y}_i\right\|_2$ ($\boldsymbol{y}_i$ is the one-hot vector of the true label). As a result, an estimate of GraND is \textbf{EL2N}, which is defined as follows:
\begin{equation}
\label{eq:el2n}
    \text{EL2N}(x_i, y_i) = \mathbb{E}_{\boldsymbol{w}}\left\|p(x_i) - \boldsymbol{y}_i\right\|_2
\end{equation}
It is worth noting that this formulation is similar to the confidence metric defined by \citet{swayamdipta-2020-dataset-cartography}. 
When computing confidence, the expected value is obtained by averaging the model output probabilities across the training epochs. However, as stated in Eq. \ref{eq:el2n}, the EL2N expectation is over multiple weight initializations ($\boldsymbol{w}$).

\section{Experiments}
In this section, we verify the effectiveness of GraNd and EL2N \citep{NEURIPS2021_ac56f8fe_datadiet} in the NLP domain. 
Our experimental setup is similar to \citet{NEURIPS2021_ac56f8fe_datadiet} for the followings:
(1) models are trained on the subsets of the data with higher GraNd and EL2N scores;
(2) based on the expectation over multiple weight initializations mentioned in Eq.~\ref{eq:grand} and Eq.~\ref{eq:el2n}, we average the scores over five independent training runs.\footnote{It is ten independent training runs in \citet{NEURIPS2021_ac56f8fe_datadiet}}; and,
(3) for putting our results in context, we utilized random pruning as the baseline\footnote{Random selection with the same proportion} and the accuracy of training on the whole dataset with no pruning as the target performance.
\begin{figure*}[t!]
\centering
    \subfloat{
        \includegraphics[width=0.38\textwidth, trim=5 0 20 10, clip] {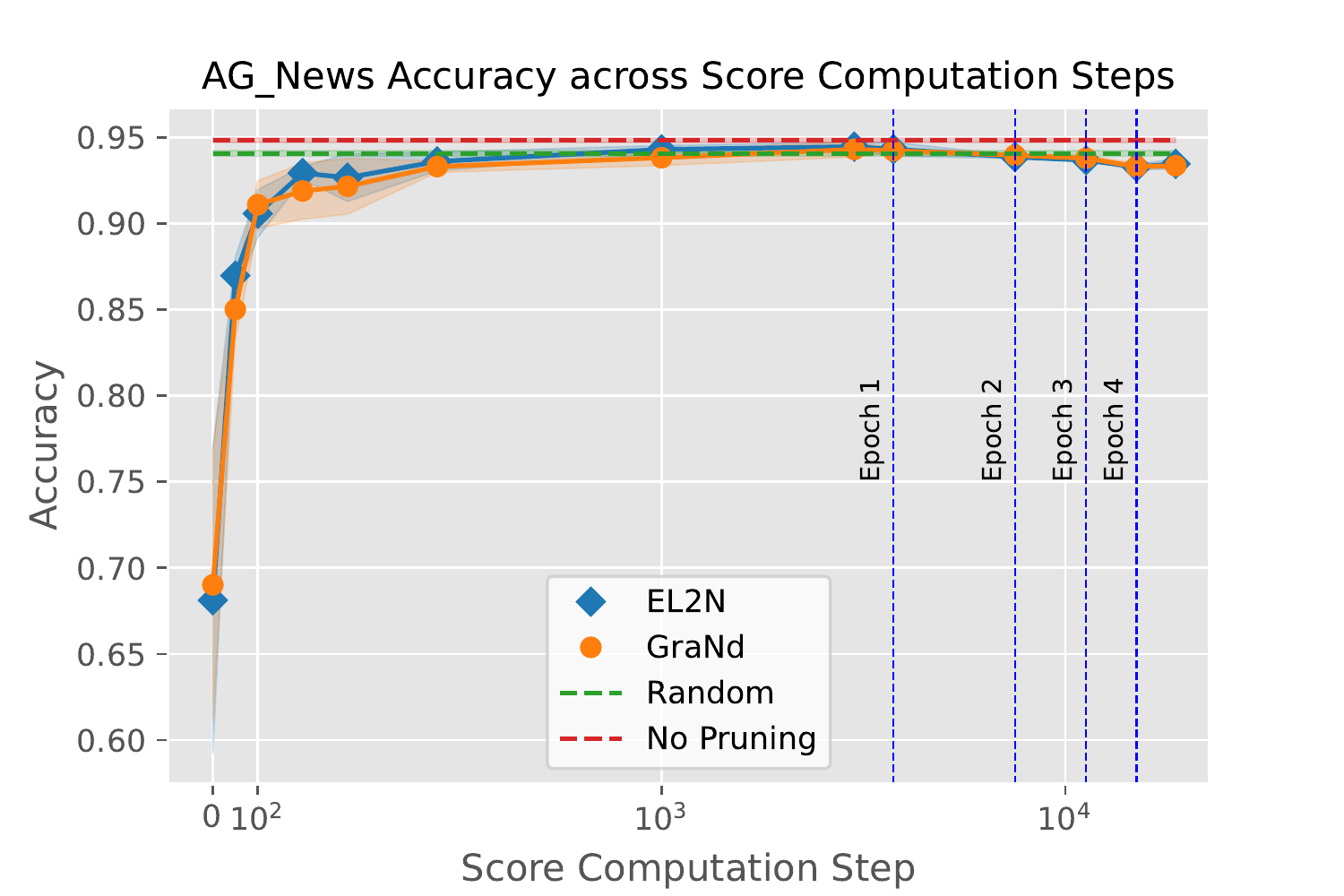}
    }
    \subfloat{
        \includegraphics[width=0.38\textwidth, trim=5 0 20 10, clip] {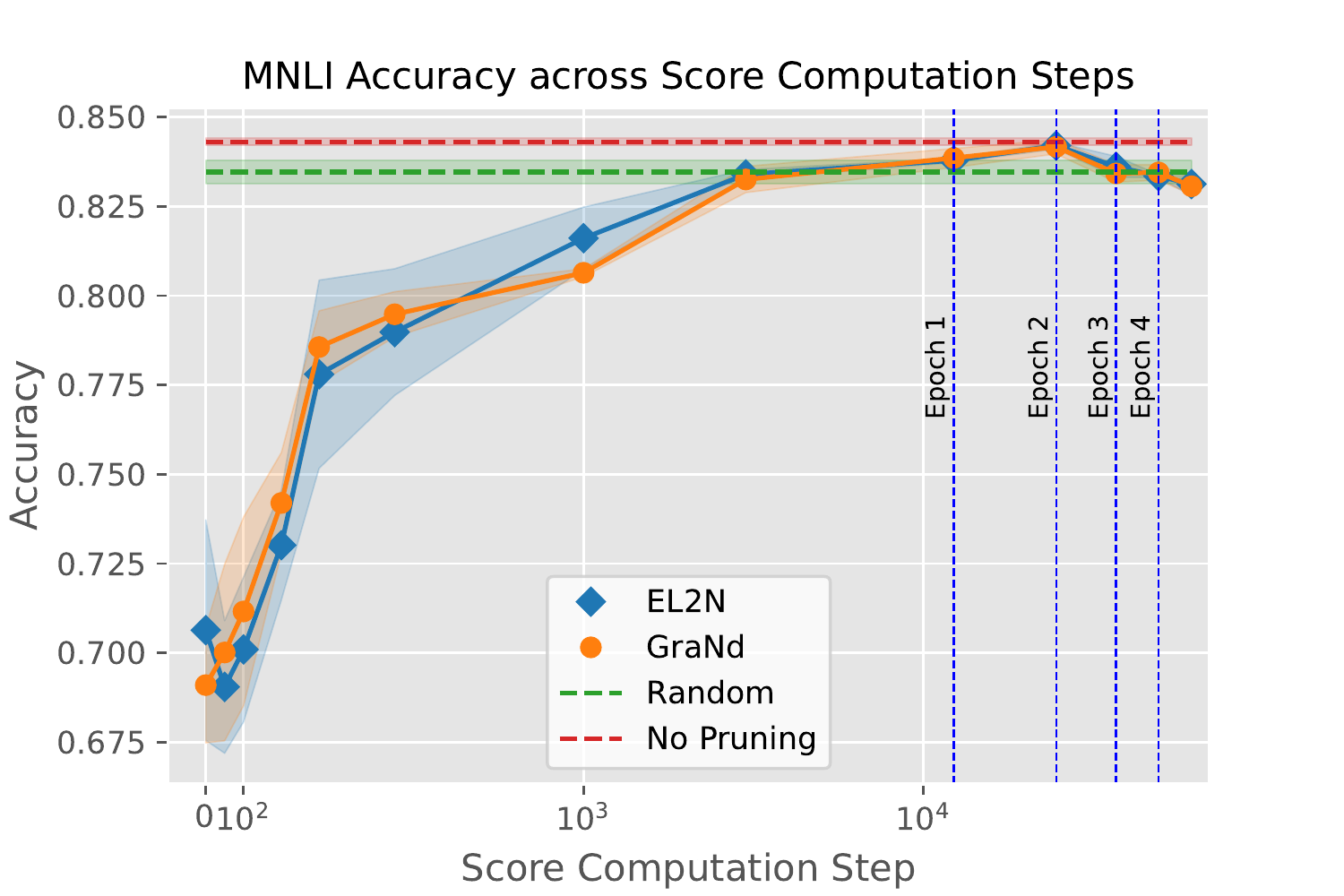}
    }

    \caption{
    MNLI/AGNews accuracy of training BERT-base on the top 70\%/50\% of examples with the highest EL2N or GraNd scores computed at various steps of fine-tuning. Each point is the average of three runs and the standard deviation is shown as the shaded area. Early score computation steps are shown to result in lower accuracies.
    }
    \label{fig:steps}
\end{figure*}
Two major differences between these two setups are mentioned here:
(1) we used a pre-trained model, i.e., BERT \citep{devlin-etal-2019-bert}, standard in the NLP domain, whereas \citet{NEURIPS2021_ac56f8fe_datadiet} uses vision models initialized with random weights; and
(2) as fine-tuning requires few epochs of training, we computed the metrics over fine-grained steps rather than epochs.

\subsection{Datasets and Setup}
\paragraph{Datasets.}
We evaluate our methods on two different classification tasks. 
For natural language inference, we used MNLI dataset \citep{williams-etal-2018-broad}, and for topic classification, we used AG's~News \citep{Zhang2015CharacterlevelCN_agnews}. 
We report the validation set (matched) and test set accuracy, respectively.

\paragraph{Setup.}
We used the Transformers library from HuggingFace \citep{wolf-etal-2020-transformers} and BERT-base-uncased as our pre-trained language model.\footnote{Our evaluations are based on the BERT's base version (12-layer, 768-hidden size, 12-attention head, 110M parameters) due to the massive number of experiments and resource limitations.} 
To fine-tune BERT, we trained the model for five epochs and selected the best performance, with 3e-5 as the learning rate. For calculating GraNd and EL2N scores, we used batch sizes of 12 and 32 respectively executed on a Tesla P100 GPU.
\begin{wrapfigure}{r}{0.33\textwidth}
\centering
    \includegraphics[width=0.31\textwidth, trim=0 0 0 5, clip] {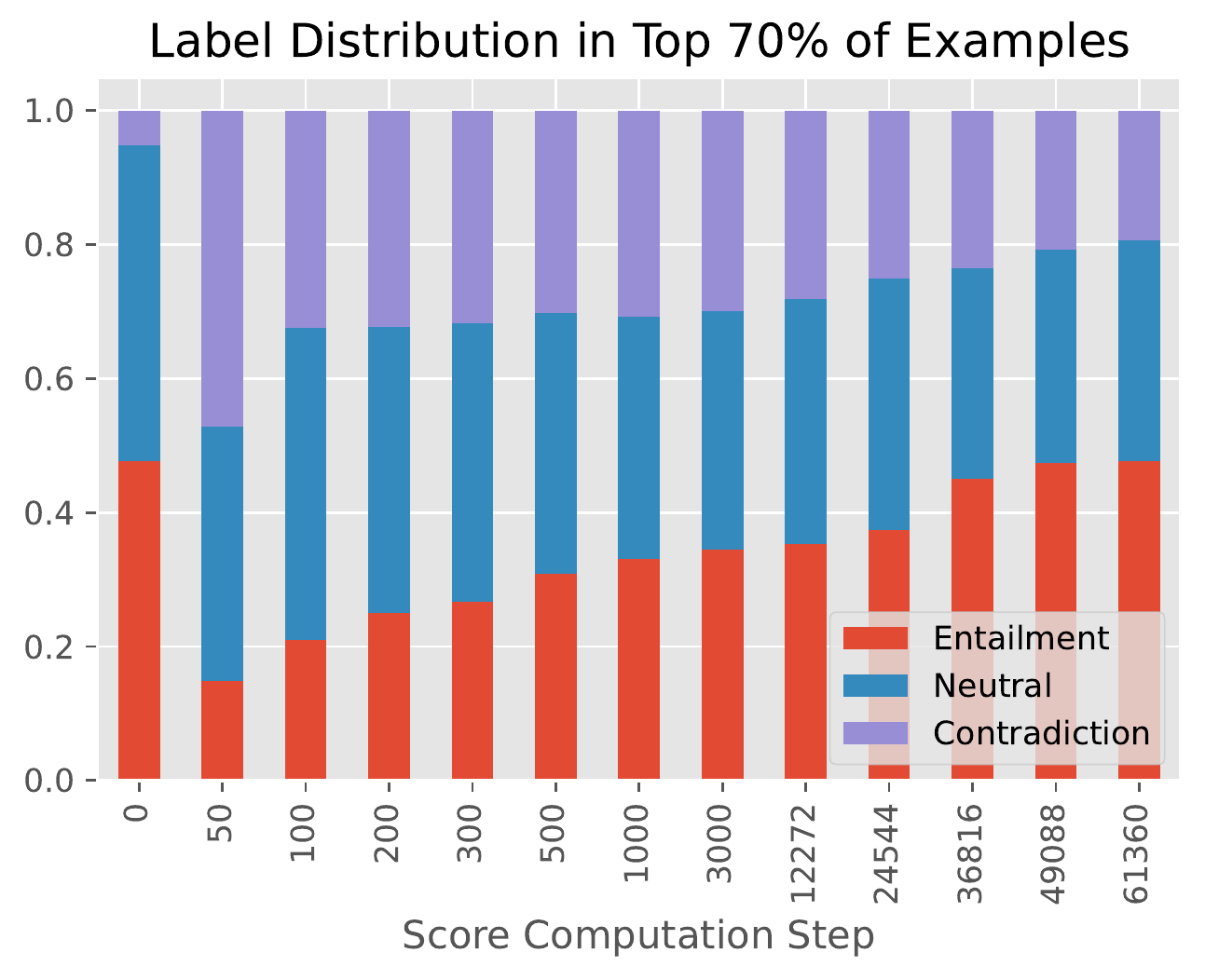}
    \caption{
    The distribution of labels of top-70\% examples with the highest EL2N scores in MNLI across score computation steps. 
    The distribution is extremely unbalanced in early steps.
    }
    \vspace{-40pt}
    \label{fig:labels}
\end{wrapfigure}


\subsection{Results}
\paragraph{Score computation step.}
Figure~\ref{fig:steps} demonstrates BERT's performance on MNLI and AG's News datasets after fine-tuning on the respective top 70\% and 50\% of examples, ranked by EL2N and GraNd scores.
Most notably, early score computation steps are shown not to be reliable for obtaining a representative subset of the dataset to the extent that they may result in a lower performance than even random sampling. 

To further investigate this, Figure~\ref{fig:labels} shows the distribution of labels in the top examples chosen by EL2N across score computation steps. 
The distribution is extremely unbalanced in early steps which can explain the low performance of fine-tuning on high-scoring examples at the early stages.
Moreover, the two datasets show different behaviours. 
AG's News seems to provide acceptable scores after only 500 steps, while MNLI takes longer to reach the random baseline.
For further experiments we used the first epoch as the score computation step which is 3,750 and 12,272 for AG's~News and MNLI, respectively.

\begin{figure*}[t]
\centering
    \subfloat{
        \includegraphics[width=0.38\textwidth, trim=0 0 40 10, clip] {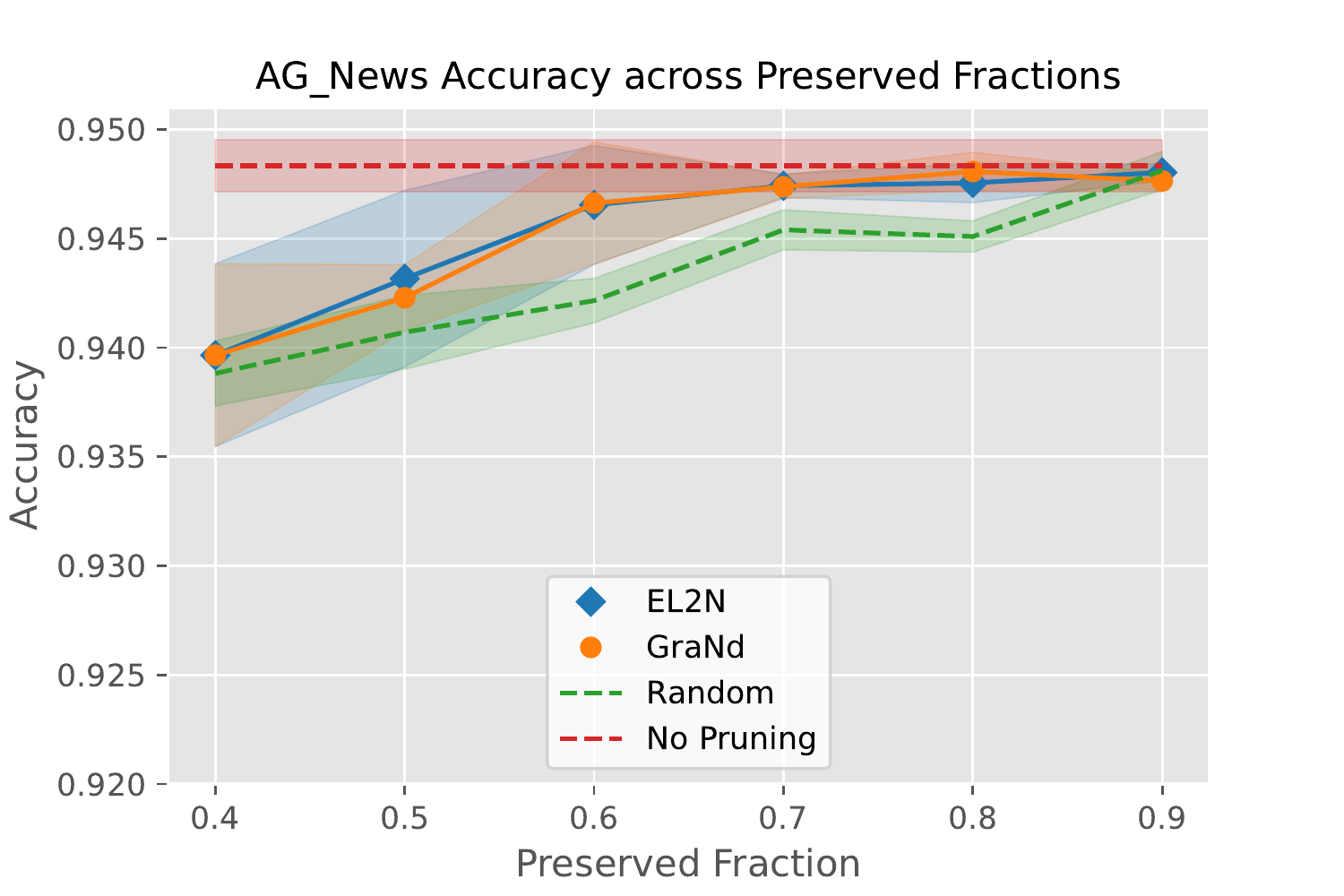}
    }
    \subfloat{
        \includegraphics[width=0.38\textwidth, trim=0 0 40 10, clip] {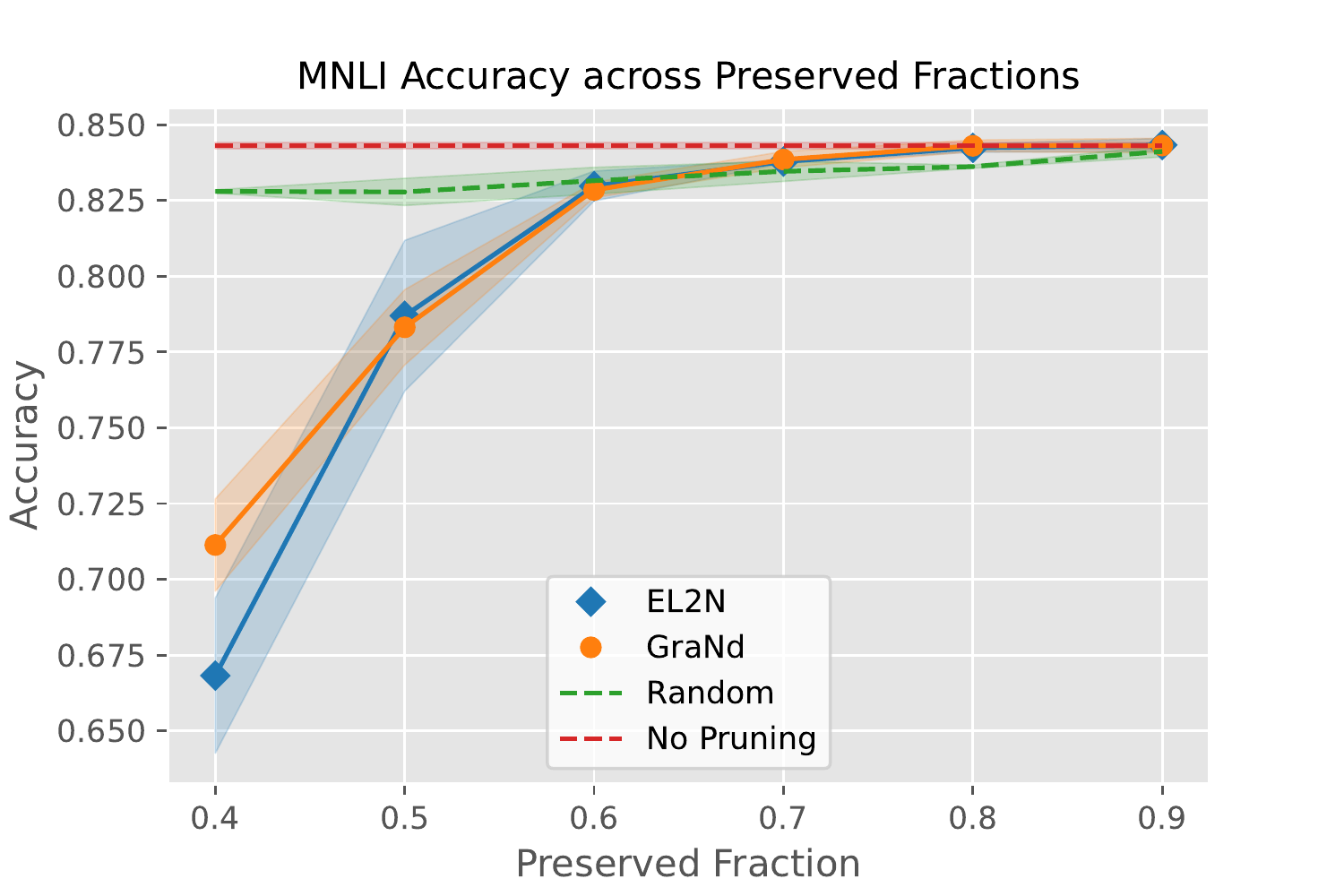}
    }

    \caption{
    MNLI/AGNews accuracy of training BERT-base on the top k\% of the examples with the highest EL2N/GraNd scores computed after one epoch of fine-tuning. Each point is the average of three runs and the standard deviation is shown as the shaded area.
    The performance grows as the preserved fraction increases.
    }
    \label{fig:fractions}
\end{figure*}
\paragraph{Preserved fraction.}
To examine how the size of training set can affect the performance, we fine-tuned the model using different percentages of the entire dataset. 
Figure~\ref{fig:fractions} shows the performance of BERT-base trained on the top k\% of the data points with the highest EL2N/GraNd scores calculated after one epoch of fine-tuning. As shown in Figure~\ref{fig:fractions}, the models fine-tuned on smaller subsets of the dataset could not perform as well as the one fine-tuned on the whole dataset in both tasks and scoring metrics. 
Albeit, we see the growing performance as the preserved fraction increases. Interestingly, in MNLI, a model trained on the chosen subset performs worse than a random subset until 60\% of data is preserved. This is contradicting with AG's News where even in small portions, EL2N and GraNd could outperform random baseline.
Nevertheless, the EL2N and GraNd scores are shown to be better than random pruning when we preserve 70\% or more of the dataset.

\begin{figure*}[b!]
\centering
    \subfloat{
        \includegraphics[width=0.38\textwidth, trim=5 0 5 5, clip] {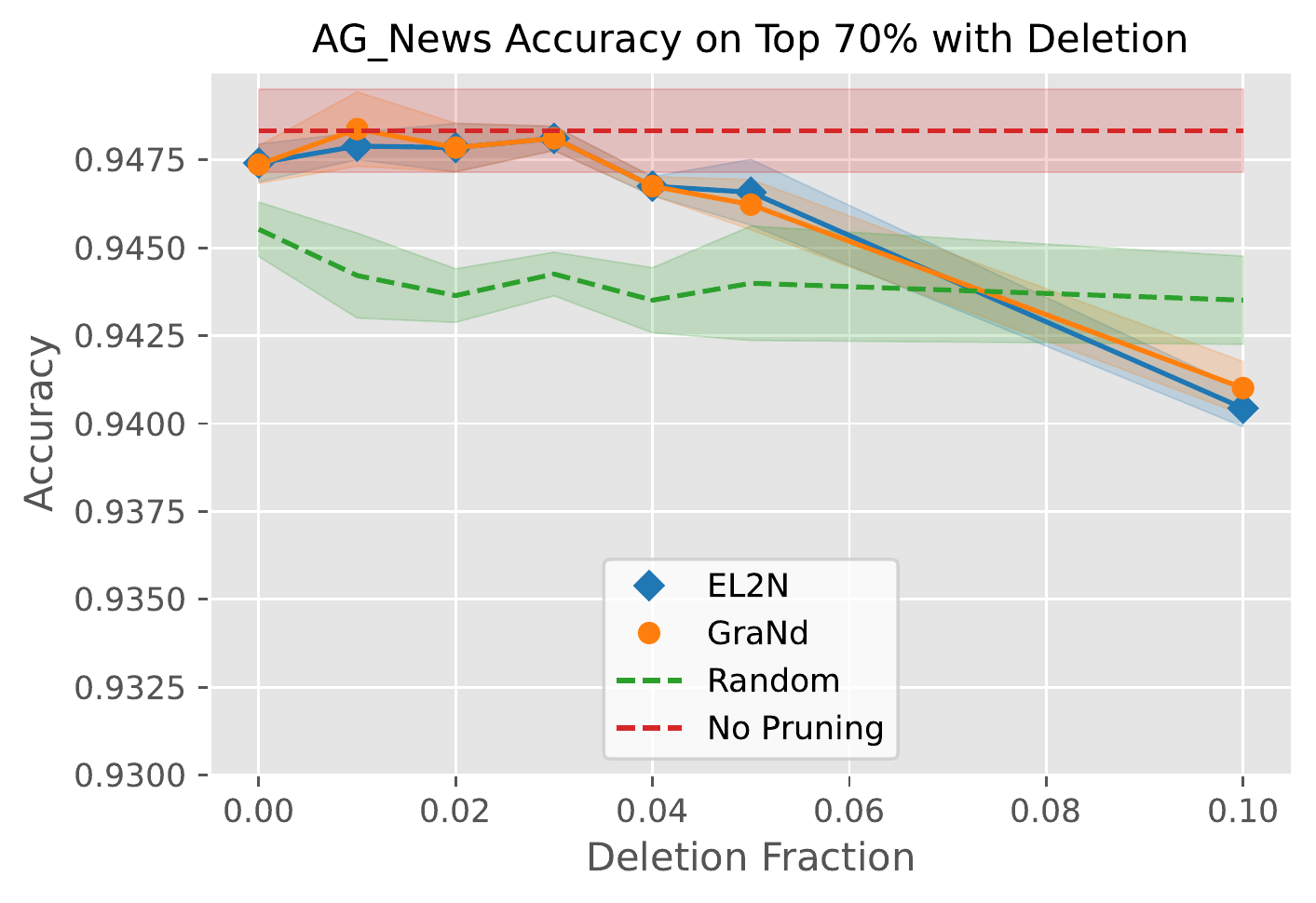}
    }
    \subfloat{
        \includegraphics[width=0.38\textwidth, trim=5 0 5 5, clip] {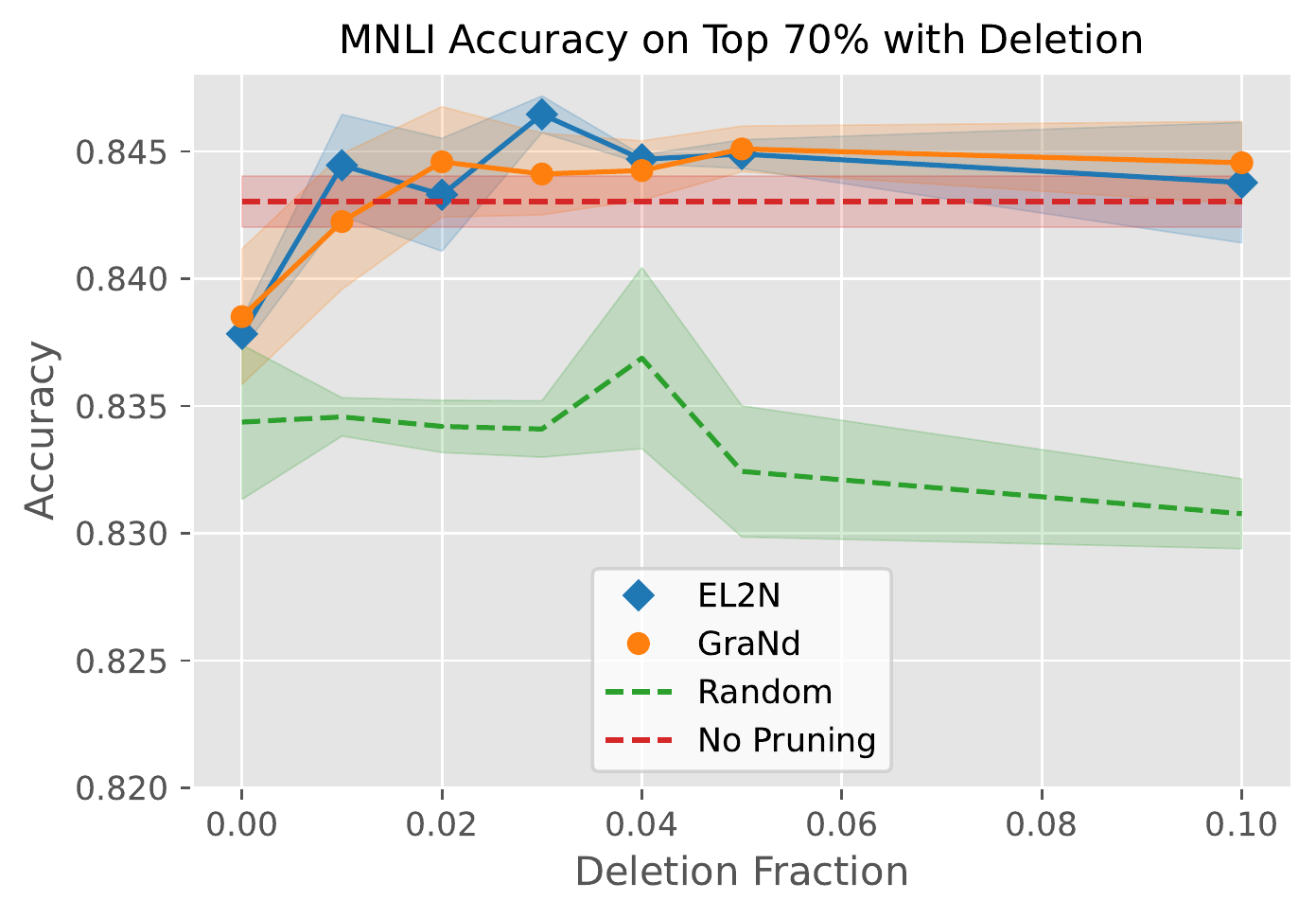}
    }

    \caption{
    MNLI/AGNews accuracy of training BERT-base on the top 70\% of examples with the highest EL2N/GraNd scores computed at first epoch of fine-tuning with top k\% deleted.
    An increase in performance is observed in both datasets when deleting a small portion of highest scoring examples.
    }
    \label{fig:deletions}
\end{figure*}
\paragraph{Noise examples.}
As discussed in recent research, datasets often include noisy examples, detection of which has been of interest \citep{swayamdipta-2020-dataset-cartography, jindal-etal-2019-effective, chen-yu-and-samuel-r-bowman-2022-clean}.
We conducted an experiment to see whether removing the highest scoring examples in EL2N and GraNd, which might correspond to noisy examples \citep{NEURIPS2021_ac56f8fe_datadiet}, can improve the final performance.
Figure~\ref{fig:deletions} illustrates the performance of BERT-base after being trained on the top 70\% of data and when the top k\% of it is removed.
An increase in performance can be seen in both datasets initially which can explain that deleting a conservative amount of the highest scoring examples can help the model to learn better.
Remarkably, in the MNLI dataset we see that removing the top scoring examples achieves even higher accuracies than training on the whole dataset. This shows that, on the MNLI dataset, it is possible to reach higher performance when training on about 66\% (70\%-4\%) than the whole 100\%.
On the AG's News, however, pruning does not result in performance gain, but comparable performance can be obtained with only 67\% of the data.


\paragraph{Computation concerns.}
For computing EL2N scores, we employed the average across five seeds, and each seed is trained for one epoch. It is computationally equivalent to total of five epochs of training. However, we argue that it is possible to use fewer seeds to achieve scores which highly correlate to the mean of five seeds. Average spearman correlation of the scores between each of those seeds and the mean of five seeds is $0.9311$ with standard deviation of $0.0042$ which shows that using only one seed can yield very similar results to five seeds. The correlation increases to $0.9722$ between average of two seeds and five seeds. This sheds light on the fact that even with limited resources, EL2N may find the important examples by only fine-tuning a few seeds, each for one epoch.


\section{Conclusions}
We adapted two data pruning metrics from computer vision, called EL2N and GraNd, to NLP. We showed that despite the major differences between the two fields, we can find subsets of data that maintain or, in some cases, even improve the performance. We demonstrated that unlike in vision, neither GraNd nor EL2N can yield acceptable performances in early steps of fine-tuning. 
In summary, at least one epoch of  fine-tuning is necessary for a reliable computation of either of the two scores. 
Finally, we explained that, despite being dataset dependent, pruning the highest scoring examples, which may be related to noise, can achieve higher accuracies than when training on the whole dataset.
A potential future work could be an \mbox{end-to-end} pruning mechanism with a single fine-tuning procedure.

\bibliography{anthology,custom}
\bibliographystyle{acl_natbib}

\appendix

\section{Appendix}

\subsection{Dataset samples and their scores}
Examples with the highest and lowest EL2N scores are provided in Table~\ref{tab:examples-agnews} and Table~\ref{tab:examples-mnli} for AG News and MNLI datasets. 
\begin{table*}[h]
\centering
\small
\begin{tabular}{lp{7.2cm}ccc}
\toprule 
  & \textbf{Instance} & \textbf{EL2N} & \textbf{Gold Label} & \textbf{Prediction}\\
\midrule 
\multirow{15}{*}{\rotatebox{90}{Noise / Hard}}
& Tom Hauck/Getty Images 1. Rapper Snoop Dogg attended practice Tuesday with No. 1 USC as a guest wide receiver. The Trojans were delighted by the star \#39;s presence, as were a group of pee-wee football players watching practice that day.
& 1.413128
& Business (2)
& Sports (1)
\\\cmidrule[0.01em](lr){2-5}
& Palestinian gunmen kidnap CNN producer GAZA CITY, Gaza Strip -- Palestinian gunmen abducted a CNN producer in Gaza City on Monday, the network said. The network said Riyadh Ali was taken away at gunpoint from a CNN van.
& 1.412653
& Sports (1)
& World (0)
\\\cmidrule[0.01em](lr){2-5}
& Names in the Game Dressed in jeans and a white shirt, the men \#39;s all-around champion in Athens appeared relaxed as he helped promote a 14-city gymnastics exhibition tour that kicks off at the Mohegan Sun Casino on Tuesday.
& 1.412590
& Business (2)
& Sports (1)
\\\midrule[0.03em]
\multirow{14}{*}{\rotatebox{90}{Easy}}
& Vikings \#39; Moss practices despite hamstring injury Eden Prairie, MN (Sports Network) - Minnesota Vikings wide receiver Randy Moss practiced on Wednesday despite nursing a strained right hamstring and is listed as questionable for this Sunday \#39;s game against the New York Giants.
& 0.000665
& Sports (1)
& Sports (1)
\\\cmidrule[0.01em](lr){2-5}
& Cassell a no-show; Wolves sign Griffin Minneapolis, MN (Sports Network) - Minnesota Timberwolves guard Sam Cassell did not show up Tuesday for the first day of training camp.
& 0.000653
& Sports (1)
& Sports (1)
\\\cmidrule[0.01em](lr){2-5}
& Rockets activate Lue from injured list Houston, TX (Sports Network) - The Houston Rockets activated guard Tyronn Lue from the injured list prior to Wednesday \#39;s game against the Hawks.	
& 0.000649
& Sports (1)
& Sports (1)
 \\
\bottomrule
\end{tabular}
\caption{
Examples from AG News belonging to different score regions of EL2N. The highest scoring instances are mostly noisy samples, while the least scoring instances are very easy to learn.
} 
\label{tab:examples-agnews}
\end{table*}
\begin{table*}[h]
\centering
\small
\begin{tabular}{lp{2.7cm}p{2.7cm}ccc}
\toprule 
  & \textbf{Premise} & \textbf{Hypothesis} & \textbf{EL2N} & \textbf{Gold Label} & \textbf{Prediction}\\
\midrule 
\multirow{10}{*}{\rotatebox{90}{Noise / Hard}}
& Social insurance taxes and contributions paid by Federal employees (575)	
& There are no taxes for the Federal employees.	
& 1.410452
& Entailment (0)
& Contradiction (2)
\\\cmidrule[0.01em](lr){2-6}
& um-hum um-hum yep you were very fortunate
& You were very unfortunate.	
& 1.410408
& Entailment (0)
& Contradiction (2)
\\\cmidrule[0.01em](lr){2-6}
& "Everyone is chanting for you," Nema told him.
& Everyone was silent.
& 1.410146
& Neutral (1)
& Contradiction (2)
\\\midrule[0.03em]
\multirow{7}{*}{\rotatebox{90}{Easy}}
& Many of them did.	
& None of them did.	
& 0.002198
& Contradiction (2)
& Contradiction (2)
\\\cmidrule[0.01em](lr){2-6}
& Yes ”doubly careful." He turned to me abruptly.	
& No, not careful at all. He slunk away.	
& 0.002147
& Contradiction (2)
& Contradiction (2)
\\\cmidrule[0.01em](lr){2-6}
& Many others exist, too.
& There are no others who exist.
& 0.002134
& Contradiction (2)
& Contradiction (2)
 \\
\bottomrule
\end{tabular}
\caption{
Examples from MNLI belonging to different score regions of EL2N. The highest scoring instances are mostly noisy samples, while the least scoring instances are very easy to learn.
} 
\label{tab:examples-mnli}
\end{table*}


\end{document}